\documentclass[11pt,a4paper]{article}
\usepackage[hyperref]{acl2018}
\usepackage{times}
\usepackage{latexsym}
\usepackage{bm}
\usepackage{amsmath,amsfonts,amssymb}
\usepackage{subcaption}
\usepackage{mathtools}

\usepackage[utf8]{inputenc}

\usepackage{booktabs}
\usepackage{multirow}
\usepackage{graphicx}
\usepackage{url}
\usepackage{todonotes}

\aclfinalcopy 
\def\aclpaperid{107} 

\DeclarePairedDelimiter\floor{\lfloor}{\rfloor}

\title{Universal Language Model Fine-tuning for Text Classification}

\author{Jeremy Howard\footnotemark \\
  fast.ai \\
  University of San Francisco\\
  {\tt j@fast.ai} \\\And
  Sebastian Ruder\footnotemark[1] \\
  Insight Centre, NUI Galway \\
  Aylien Ltd., Dublin \\
  {\tt sebastian@ruder.io} \\}

\date{}

\begin{document}
\maketitle 
\begin{abstract}
Inductive transfer learning has greatly impacted computer vision, but existing approaches in NLP still require task-specific modifications and training from scratch. We propose Universal Language Model Fine-tuning (ULMFiT), an effective transfer learning method that can be applied to any task in NLP, and introduce techniques that are key for fine-tuning a language model. Our method significantly outperforms the state-of-the-art on six text classification tasks, reducing the error by 18-24\% on the majority of datasets. Furthermore, with only $100$ labeled examples, it matches the performance of training from scratch on $100\times$ more data. We open-source our pretrained models and code\footnote{\url{http://nlp.fast.ai/ulmfit}.}.
\end{abstract}

\newenvironment{starfootnotes}
  {\par\edef\savedfootnotenumber{\number\value{footnote}}
   \renewcommand{\thefootnote}{$\star$} 
   \setcounter{footnote}{0}}
  {\par\setcounter{footnote}{\savedfootnotenumber}}

\begin{starfootnotes}
\footnotetext{Equal contribution. Jeremy focused on the algorithm development and implementation, Sebastian focused on the experiments and writing.}
\end{starfootnotes}

\section{Introduction}

Inductive transfer learning has had a large impact on computer vision (CV). Applied CV models (including object detection, classification, and segmentation) are rarely trained from scratch, but instead are fine-tuned from models that have been pretrained on ImageNet, MS-COCO, and other datasets \cite{sharif2014cnn,long2015fully,He2015,Huang2017}.

Text classification is a category of Natural Language Processing (NLP) tasks with real-world applications such as spam, fraud, and bot detection \cite{jindal2007review,ngai2011application,chu2012detecting}, emergency response \cite{caragea2011classifying}, and commercial document classification, such as for legal discovery \cite{roitblat2010document}.

While Deep Learning models have achieved state-of-the-art on many NLP tasks, these models are trained from scratch, requiring large datasets, and days to converge. Research in NLP focused mostly on \emph{transductive} transfer \cite{Blitzer2007}. For \emph{inductive} transfer, fine-tuning pretrained word embeddings \cite{Mikolov2013d}, a simple transfer technique that only targets a model's first layer, has had a large impact in practice and is used in most state-of-the-art models. Recent approaches that concatenate embeddings derived from other tasks with the input at different layers \cite{peters2017semi,Mccann2017,deepcontext2017} still train the main task model from scratch and treat pretrained embeddings as fixed parameters, limiting their usefulness. 

In light of the benefits of pretraining \cite{erhan2010does}, we should be able to do better than \emph{randomly initializing} the remaining parameters of our models. However, inductive transfer via fine-tuning has been unsuccessful for NLP \cite{Mou2016}. \newcite{Dai2015} first proposed fine-tuning a language model (LM) but require millions of in-domain documents to achieve good performance, which severely limits its applicability.

We show that not the idea of LM fine-tuning but our lack of knowledge of how to train them effectively has been hindering wider adoption. LMs overfit to small datasets and suffered catastrophic forgetting when fine-tuned with a classifier. Compared to CV, NLP models are typically more shallow and thus require different fine-tuning methods. 

We propose a new method, Universal Language Model Fine-tuning (ULMFiT) that addresses these issues and enables robust inductive transfer learning for any NLP task, akin to fine-tuning ImageNet models: The same 3-layer LSTM architecture---with the same hyperparameters and no additions other than tuned dropout hyperparameters---outperforms highly engineered models and transfer learning approaches on six widely studied text classification tasks. On IMDb, with $100$ labeled examples, ULMFiT matches the performance of training from scratch with $10\times$ and---given $50$k unlabeled examples---with $100\times$ more data.

\paragraph{Contributions} Our contributions are the following: 1) We propose Universal Language Model Fine-tuning (ULMFiT), a method that can be used to achieve CV-like transfer learning for any task for NLP. 2) We propose \emph{discriminative fine-tuning}, \emph{slanted triangular learning rates}, and \emph{gradual unfreezing}, novel techniques to retain previous knowledge and avoid catastrophic forgetting during fine-tuning. 3) We significantly outperform the state-of-the-art on six representative text classification datasets, with an error reduction of 18-24\% on the majority of datasets. 4) We show that our method enables extremely sample-efficient transfer learning and perform an extensive ablation analysis. 5) We make the pretrained models and our code available to enable wider adoption.

\section{Related work}

\paragraph{Transfer learning in CV} Features in deep neural networks in CV have been observed to transition from \emph{general} to task-\emph{specific} from the first to the last layer \cite{yosinski2014transferable}. For this reason, most work in CV focuses on transferring the first layers of the model \cite{Long2015a}. \newcite{sharif2014cnn} achieve state-of-the-art results using features of an ImageNet model as input to a simple classifier. In recent years, this approach has been superseded by fine-tuning either the last \cite{donahue2014decaf} or several of the last layers of a pretrained model and leaving the remaining layers frozen \cite{long2015fully}.

\paragraph{Hypercolumns} In NLP, only recently have methods been proposed that go beyond transferring word embeddings. The prevailing approach is to pretrain embeddings that capture additional context via other tasks. Embeddings at different levels are then used as features, concatenated either with the word embeddings or with the inputs at intermediate layers. This method is known as hypercolumns \cite{hariharan2015hypercolumns} in CV\footnote{A hypercolumn at a pixel in CV is the vector of activations of all CNN units above that pixel. In analogy, a hypercolumn for a word or sentence in NLP is the concatenation of embeddings at different layers in a pretrained model.} and is used by \newcite{peters2017semi}, \newcite{deepcontext2017}, \newcite{Wieting2017}, \newcite{Conneau2017}, and \newcite{Mccann2017} who use language modeling, paraphrasing, entailment, and Machine Translation (MT) respectively for pretraining. Specifically, \newcite{deepcontext2017} require engineered custom architectures, while we show state-of-the-art performance with the same basic architecture across a range of tasks. In CV, hypercolumns have been nearly entirely superseded by end-to-end fine-tuning \cite{long2015fully}.

\paragraph{Multi-task learning} A related direction is multi-task learning (MTL) \cite{Caruana1993}. This is the approach taken by \newcite{semisupervised2017rei} and \newcite{empower2018liu} who add a language modeling objective to the model that is trained jointly with the main task model.
MTL requires the tasks to be trained from scratch every time, which makes it inefficient and often requires careful weighting of the task-specific objective functions \cite{Chen2017}.
\paragraph{Fine-tuning} Fine-tuning has been used successfully to transfer between similar tasks, e.g. in QA \cite{Min2017}, for distantly supervised sentiment analysis \cite{Severyn2015a}, or MT domains \cite{sennrich2015improving} but has been shown to fail between unrelated ones \cite{Mou2016}. \newcite{Dai2015} also fine-tune a language model, but overfit with $10$k labeled examples and require millions of in-domain documents for good performance. In contrast, ULMFiT leverages general-domain pretraining and novel fine-tuning techniques to prevent overfitting even with only $100$ labeled examples and achieves state-of-the-art results also on small datasets.

\begin{figure*}[!htb]
    \begin{subfigure}{.30\linewidth}
      \centering
         \includegraphics[height=2.2in]{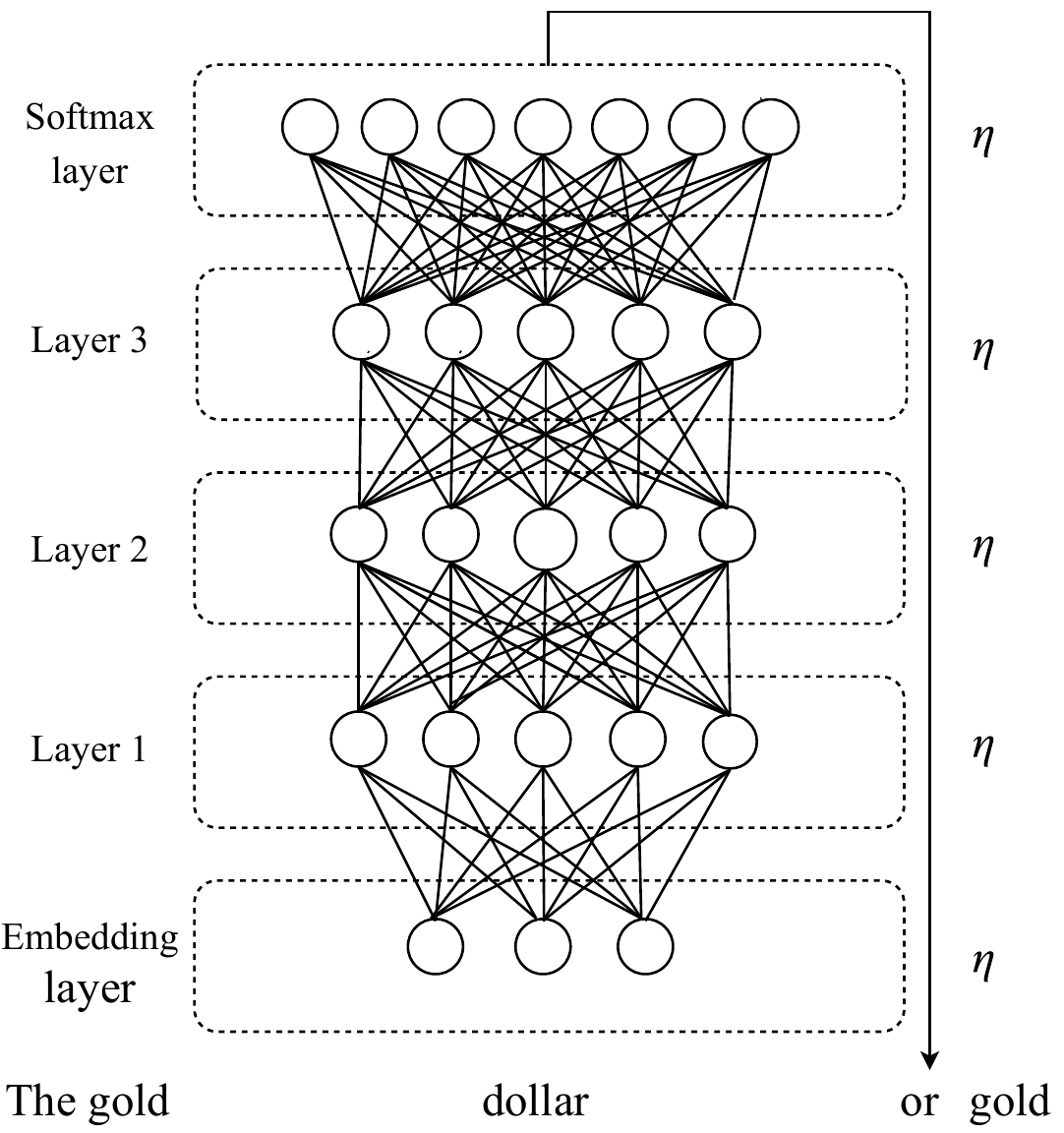}
    \caption{LM pre-training} \label{fig:lm-pretraining}
    \end{subfigure}%
    \hspace*{0.4cm}
    \begin{subfigure}{.30\linewidth}
      \centering
         \includegraphics[height=2.2in]{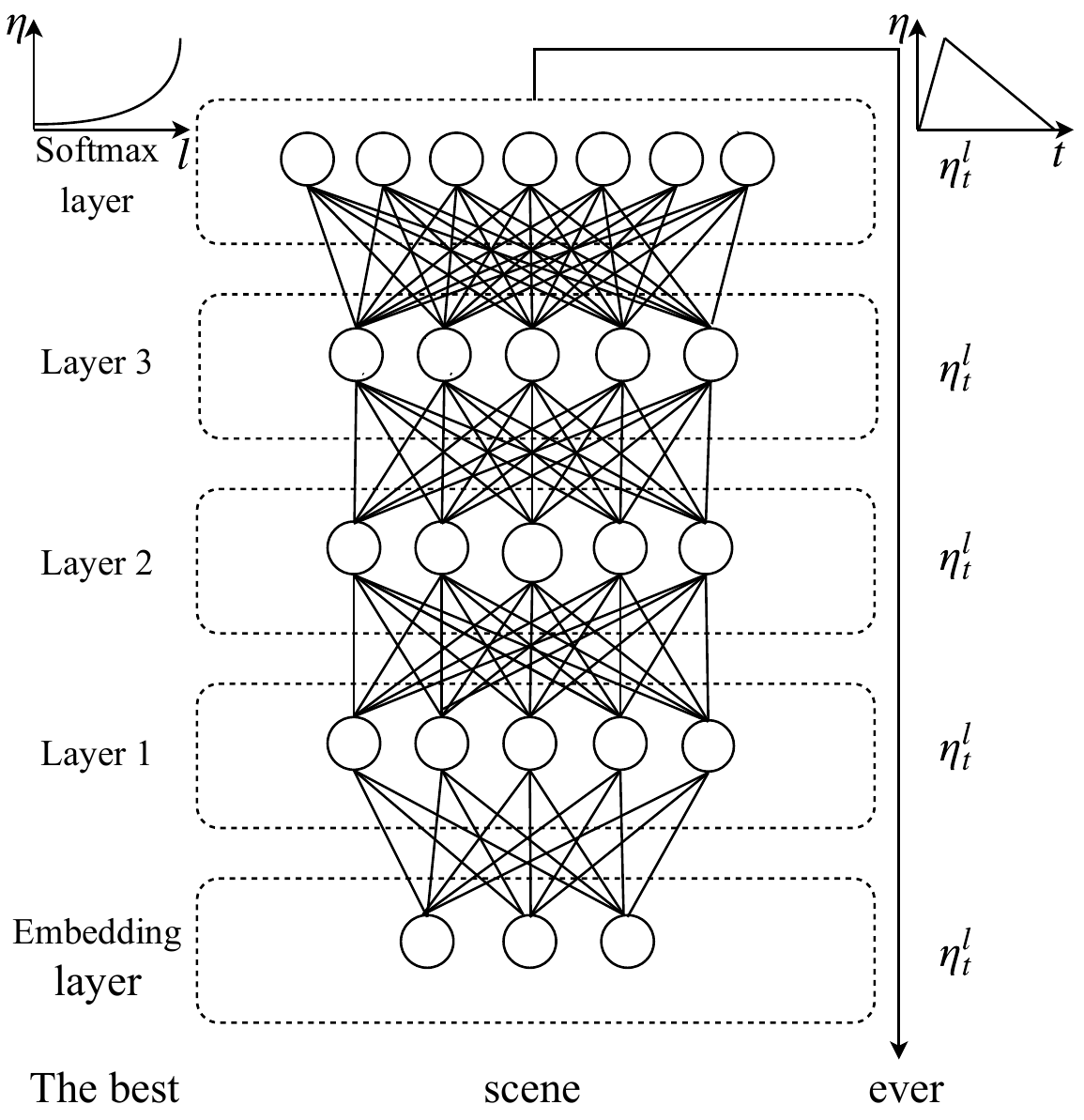}
    \caption{LM fine-tuning} \label{fig:lm-fine-tuning}
    \end{subfigure}
    \hspace*{0.4cm}
    \begin{subfigure}{.30\linewidth}
      \centering
         \includegraphics[height=2.2in]{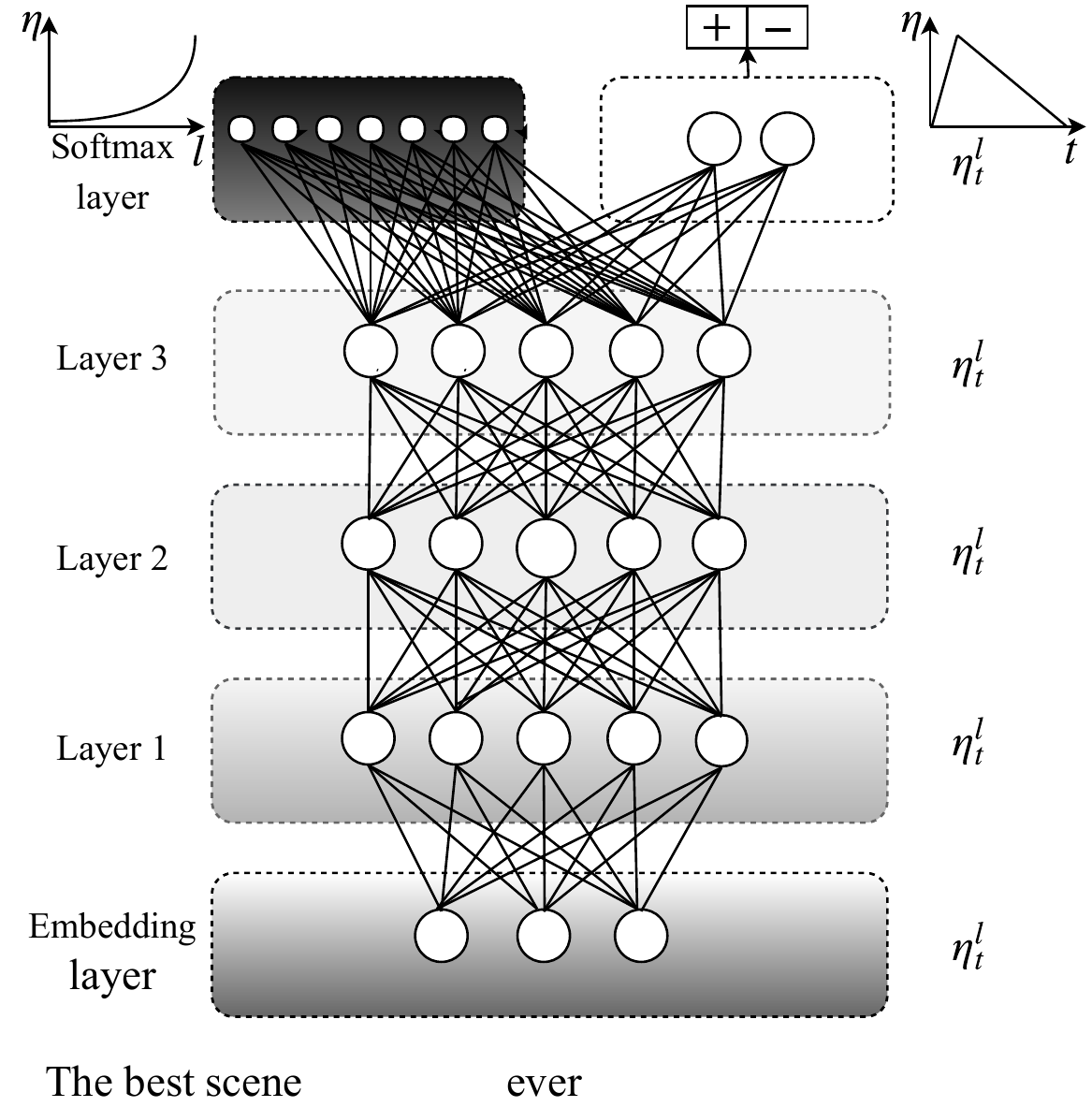}
    \caption{Classifier fine-tuning} \label{fig:classifier-fine-tuning}
    \end{subfigure}
    \caption{ULMFiT consists of three stages: a) The LM is trained on a general-domain corpus to capture general features of the language in different layers. b) The full LM is fine-tuned on target task data using discriminative fine-tuning (`\emph{Discr}') and slanted triangular learning rates (STLR) to learn task-specific features. c) The classifier is fine-tuned on the target task using gradual unfreezing, `\emph{Discr}', and STLR to preserve low-level representations and adapt high-level ones (shaded: unfreezing stages; black: frozen).}
\label{fig:ulmfit}
\end{figure*}

\section{Universal Language Model Fine-tuning}

We are interested in the most general \emph{inductive} transfer learning setting for NLP \cite{Pan2010}: Given a static source task $\mathcal{T}_S$ and \emph{any} target task $\mathcal{T}_T$ with $\mathcal{T}_S \neq \mathcal{T}_T$, we would like to improve performance on $\mathcal{T}_T$. Language modeling can be seen as the ideal source task and a counterpart of ImageNet for NLP: It captures many facets of language relevant for downstream tasks, such as long-term dependencies \cite{linzen2016assessing}, hierarchical relations \cite{Gulordava2018}, and sentiment \cite{radford2017learning}. In contrast to tasks like MT \cite{Mccann2017} and entailment \cite{Conneau2017}, it provides data in near-unlimited quantities for most domains and languages. Additionally, a pretrained LM can be easily adapted to the idiosyncrasies of a target task, which we show significantly improves performance (see Section \ref{sec:analysis}). Moreover, language modeling already is a key component of existing tasks such as MT and dialogue modeling. Formally, language modeling induces a hypothesis space $\mathcal{H}$ that should be useful for many other NLP tasks \cite{vapnik1982estimation,Baxter2000}. 

We propose Universal Language Model Fine-tuning (ULMFiT), which pretrains a language model (LM) on a large general-domain corpus and fine-tunes it on the target task using novel techniques. The method is \emph{universal} in the sense that it meets these practical criteria: 1) It works across tasks varying in document size, number, and label type; 2) it uses a single architecture and training process; 3) it requires no custom feature engineering or preprocessing; and 4) it does not require additional in-domain documents or labels.

In our experiments, we use the state-of-the-art language model AWD-LSTM \cite{Merity2017}, a regular LSTM (with no attention, short-cut connections, or other sophisticated additions) with various tuned dropout hyperparameters. Analogous to CV, we expect that downstream performance can be improved by using higher-performance language models in the future.

ULMFiT consists of the following steps, which we show in Figure \ref{fig:ulmfit}: a) General-domain LM pretraining (\textsection \ref{sec:pretraining}); b) target task LM fine-tuning (\textsection \ref{sec:lm-fine-tuning}); and c) target task classifier fine-tuning (\textsection \ref{sec:clas-fine-tuning}). We discuss these in the following sections.

\subsection{General-domain LM pretraining} \label{sec:pretraining}

An ImageNet-like corpus for language should be large and capture general properties of language. We pretrain the language model on Wikitext-103 \cite{Merity2016} consisting of 28,595 preprocessed Wikipedia articles and 103 million words. Pretraining is most beneficial for tasks with small datasets and enables generalization even with $100$ labeled examples. We leave the exploration of more diverse pretraining corpora to future work, but expect that they would boost performance. While this stage is the most expensive, it only needs to be performed once and improves performance and convergence of downstream models.

\subsection{Target task LM fine-tuning} \label{sec:lm-fine-tuning}

No matter how diverse the general-domain data used for pretraining is, the data of the target task will likely come from a different distribution. We thus fine-tune the LM on data of the target task. Given a pretrained general-domain LM, this stage converges faster as it only needs to adapt to the idiosyncrasies of the target data, and it allows us to train a robust LM even for small datasets. We propose \emph{discriminative fine-tuning } and \emph{slanted triangular learning rates} for fine-tuning the LM, which we introduce in the following.

\paragraph{Discriminative fine-tuning}

As different layers capture \emph{different types of information} \cite{yosinski2014transferable}, they should be fine-tuned to \emph{different extents}.
To this end, we propose a novel fine-tuning method, \emph{discriminative fine-tuning}\footnote{
An unrelated method of the same name exists for deep Boltzmann machines \cite{salakhutdinov2009deep}.}.

Instead of using the same learning rate for \emph{all} layers of the model, discriminative fine-tuning allows us to tune \emph{each} layer with different learning rates. For context, the regular stochastic gradient descent (SGD) update of a model's parameters $\theta$ at time step $t$ looks like the following \cite{ruder2016overview}:
\begin{equation}
\theta_{t} = \theta_{t-1} - \eta \cdot \nabla_\theta J(\theta)
\end{equation}
where $\eta$ is the learning rate and $\nabla_\theta J(\theta)$ is the gradient with regard to the model's objective function. For discriminative fine-tuning, we split the parameters $\theta$ into $\{\theta^1, \ldots, \theta^L \}$ where $\theta^l$ contains the parameters of the model at the $l$-th layer and $L$ is the number of layers of the model. Similarly, we obtain $\{\eta^1, \ldots, \eta^L \}$ where $\eta^l$ is the learning rate of the $l$-th layer.

The SGD update with discriminative fine-tuning is then the following:
\begin{equation}
\theta_{t}^l = \theta_{t-1}^l - \eta^l \cdot \nabla_{\theta^l} J(\theta)
\end{equation}
We empirically found it to work well to first choose the learning rate $\eta^L$ of the last layer by fine-tuning only the last layer and using $\eta^{l-1} = \eta^l / 2.6 $ as the learning rate for lower layers.

\begin{figure}[h]
\centering
\includegraphics[width=1.0\linewidth]{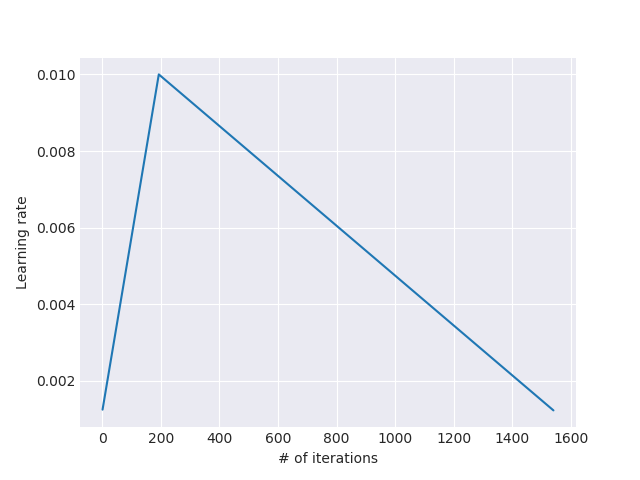}
\caption{The slanted triangular learning rate schedule used for ULMFiT as a function of the number of training iterations.}
\label{fig:triangular_lr}
\end{figure}

\paragraph{Slanted triangular learning rates} For adapting its parameters to task-specific features, we would like the model to quickly converge to a suitable region of the parameter space in the beginning of training and then refine its parameters. Using the same learning rate (LR) or an annealed learning rate throughout training is not the best way to achieve this behaviour. 
Instead, we propose \emph{slanted triangular learning rates} (STLR), which first linearly increases the learning rate and then linearly decays it according to the following update schedule, which can be seen in Figure \ref{fig:triangular_lr}:
\begin{equation}
\begin{split}
  cut & = \floor{T \cdot cut\_frac} \\
  p & =
  \begin{cases}
      t/cut, & \text{if}\ t < cut \\
      1 - \frac{t-cut}{cut \cdot (1/cut\_frac - 1)}, & \text{otherwise}
  \end{cases}\\
  \eta_t & = \eta_{max} \cdot \frac{1 + p \cdot (ratio - 1)}{ratio}
\end{split}
\end{equation}
where $T$ is the number of training iterations\footnote{In other words, the number of epochs times the number of updates per epoch.}, $cut\_frac$ is the fraction of iterations we increase the LR, $cut$ is the iteration when we switch from increasing to decreasing the LR, $p$ is the fraction of the number of iterations we have increased or will decrease the LR respectively, $ratio$ specifies how much smaller the lowest LR is from the maximum LR $\eta_{max}$, and $\eta_t$ is the learning rate at iteration $t$. We generally use $cut\_frac=0.1$, $ratio=32$ and $\eta_{max} = 0.01$.

STLR modifies triangular learning rates \cite{smith2017cyclical} with a short increase and a long decay period, which we found key for good performance.\footnote{We also credit personal communication with the author.} In Section \ref{sec:analysis}, we compare against aggressive cosine annealing, a similar schedule that has recently been used to achieve state-of-the-art performance in CV \cite{Loshchilov2017}.\footnote{While \citet{Loshchilov2017} use multiple annealing cycles, we generally found one cycle to work best.}

\subsection{Target task classifier fine-tuning} \label{sec:clas-fine-tuning}

Finally, for fine-tuning the classifier, we augment the pretrained language model with two additional linear blocks. 
Following standard practice for CV classifiers, each block uses batch normalization \cite{ioffe2015batch} and dropout, with ReLU activations for the intermediate layer and a softmax activation that outputs a probability distribution over target classes at the last layer. Note that the parameters in these task-specific classifier layers are the only ones that are learned from scratch. The first linear layer takes as the input the pooled last hidden layer states.

\paragraph{Concat pooling} The signal in text classification tasks is often contained in a few words, which may occur anywhere in the document. As input documents can consist of hundreds of words, information may get lost if we only consider the last hidden state of the model. For this reason, we concatenate the hidden state at the last time step $\mathbf{h}_T$ of the document with both the max-pooled and the mean-pooled representation of the hidden states over as many time steps as fit in GPU memory $\mathbf{H} = \{\mathbf{h}_1, \ldots, \mathbf{h}_T\}$:
\begin{equation}
\mathbf{h}_c = [\mathbf{h}_T, \mathtt{maxpool}(\mathbf{H}), \mathtt{meanpool}(\mathbf{H})]
\end{equation}
where $[]$ is concatenation.

Fine-tuning the target classifier is the most critical part of the transfer learning method. Overly aggressive fine-tuning will cause catastrophic forgetting, eliminating the benefit of the information captured through language modeling; too cautious fine-tuning will lead to slow convergence (and resultant overfitting). Besides discriminative fine-tuning and triangular learning rates, we propose \emph{gradual unfreezing} for fine-tuning the classifier.

\paragraph{Gradual unfreezing} Rather than fine-tuning all layers at once, which risks catastrophic forgetting, we propose to gradually unfreeze the model starting from the last layer as this contains the \emph{least general} knowledge \cite{yosinski2014transferable}: We first unfreeze the last layer and fine-tune all unfrozen layers for one epoch. We then unfreeze the next lower frozen layer and repeat, until we fine-tune all layers until convergence at the last iteration. This is similar to `\emph{chain-thaw}' \cite{Felbo2017}, except that we add a layer at a time to the set of `thawed' layers, rather than only training a single layer at a time. 

While discriminative fine-tuning, slanted triangular learning rates, and gradual unfreezing all are beneficial on their own, we show in Section \ref{sec:analysis} that they complement each other and enable our method to perform well across diverse datasets.

\begin{table}
  \centering
\begin{tabular}{l c c c}
    \toprule
Dataset & Type & \# classes & \# examples\\
\midrule
TREC-6  & Question & 6 & 5.5k  \\
IMDb    & Sentiment & 2 & 25k   \\
Yelp-bi & Sentiment & 2 & 560k \\
Yelp-full & Sentiment & 5 & 650k \\
AG & Topic & 4 & 120k\\
DBpedia & Topic & 14 & 560k\\
\bottomrule
  \end{tabular}
      \caption{Text classification datasets and tasks with number of classes and training examples.}
  \label{tab:tasks}
\end{table}

\paragraph{BPTT for Text Classification (BPT3C)} Language models are trained with backpropagation through time (BPTT) to enable gradient propagation for large input sequences. In order to make fine-tuning a classifier for large documents feasible, we propose BPTT for Text Classification (BPT3C): We divide the document into fixed-length batches of size $b$. At the beginning of each batch, the model is initialized with the final state of the previous batch; we keep track of the hidden states for mean and max-pooling; gradients are back-propagated to the batches whose hidden states contributed to the final prediction. In practice, we use variable length backpropagation sequences \cite{Merity2017}.

\paragraph{Bidirectional language model} Similar to existing work \cite{peters2017semi,deepcontext2017}, we are not limited to fine-tuning a unidirectional language model. For all our experiments, we pretrain both a forward and a backward LM. We fine-tune a classifier for each LM independently using BPT3C and average the classifier predictions.

\section{Experiments} \label{sec:experiments}

While our approach is equally applicable to sequence labeling tasks, we focus on text classification tasks in this work due to their important real-world applications.

\begin{table*}
  \centering
  \setlength\tabcolsep{3.65pt}
\begin{tabular}{llcllc}
    \toprule
 & Model & Test &  & Model & Test\\
\midrule
 \parbox[t]{2mm}{\multirow{4}{*}{\rotatebox[origin=c]{90}{IMDb}}} & CoVe \cite{Mccann2017} & 8.2 & \parbox[t]{2mm}{\multirow{4}{*}{\rotatebox[origin=c]{90}{TREC-6}}} & CoVe \cite{Mccann2017} & 4.2 \\
 & oh-LSTM~\cite{johnson2016supervised} & 5.9 & & TBCNN~\cite{mou2015discriminative}& 4.0 \\
 & Virtual~\cite{miyato2016adversarial} & 5.9 & & LSTM-CNN~\cite{zhou2016text} & 3.9 \\
 & ULMFiT (ours) & \textbf{4.6} & & ULMFiT (ours) & \textbf{3.6} \\
\bottomrule
  \end{tabular}
      \caption{Test error rates (\%) on two text classification datasets used by \newcite{Mccann2017}.}
  \label{tab:results-mccann}
\end{table*}

\begin{table*}
\centering
\begin{tabular}{l c c c c c c c c}
\toprule
 & AG & DBpedia & Yelp-bi & Yelp-full \\
 \midrule
Char-level CNN \cite{zhang2015character} & 9.51 & 1.55 & 4.88 & 37.95 \\
CNN \cite{johnson2016supervised} & 6.57 & 0.84 & 2.90 & 32.39 \\
DPCNN \cite{johnson2017deep} & 6.87 & 0.88 & 2.64 & 30.58 \\
ULMFiT (ours) & \textbf{5.01} & \textbf{0.80} & \textbf{2.16} & \textbf{29.98} & \\
\bottomrule
\end{tabular}
\caption{Test error rates (\%) on text classification datasets used by \newcite{johnson2017deep}.}
\label{tab:results-zhang}
\end{table*}

\subsection{Experimental setup}

\paragraph{Datasets and tasks} We evaluate our method on six widely-studied datasets, with varying numbers of documents and varying document length, used by state-of-the-art text classification and transfer learning approaches \cite{johnson2017deep,Mccann2017} as instances of three common text classification tasks: sentiment analysis, question classification, and topic classification. We show the statistics for each dataset and task in Table \ref{tab:tasks}.

\paragraph{Sentiment Analysis} For sentiment analysis, we evaluate our approach on the binary movie review IMDb dataset~\cite{maas2011learning} and on the binary and five-class version of the Yelp review dataset compiled by \newcite{zhang2015character}.

\paragraph{Question Classification} We use the six-class version of the small TREC dataset~\cite{voorhees1999trec} dataset of open-domain, fact-based questions divided into broad semantic categories.

\paragraph{Topic classification} For topic classification, we evaluate on the large-scale AG news and DBpedia ontology datasets created by \newcite{zhang2015character}.

\paragraph{Pre-processing} We use the same pre-processing as in earlier work \cite{johnson2017deep,Mccann2017}. In addition, to allow the language model to capture aspects that might be relevant for classification, we add special tokens for upper-case words, elongation, and repetition.

\paragraph{Hyperparameters} We are interested in a model that performs robustly across a diverse set of tasks. To this end, if not mentioned otherwise, we use the same set of hyperparameters across tasks, which we tune on the IMDb validation set. We use the AWD-LSTM language model \cite{Merity2017} with an embedding size of $400$, $3$ layers, $1150$ hidden activations per layer, and a BPTT batch size of $70$. We apply dropout of $0.4$ to layers, $0.3$ to RNN layers, $0.4$ to input embedding layers, $0.05$ to embedding layers, and weight dropout of $0.5$ to the RNN hidden-to-hidden matrix. The classifier has a hidden layer of size $50$. We use Adam with $\beta_1=0.7$ instead of the default $\beta_1=0.9$ and $\beta_2 = 0.99$, similar to \cite{Dozat2017}. We use a batch size of $64$, a base learning rate of $0.004$ and $0.01$ for fine-tuning the LM and the classifier respectively, and tune the number of epochs on the validation set of each task\footnote{On small datasets such as TREC-6, we fine-tune the LM only for $15$ epochs without overfitting, while we can fine-tune longer on larger datasets. We found $50$ epochs to be a good default for fine-tuning the classifier.}. We otherwise use the same practices used in \cite{Merity2017}.

\paragraph{Baselines and comparison models} For each task, we compare against the current state-of-the-art. For the IMDb and TREC-6 datasets, we compare against CoVe \cite{Mccann2017}, a state-of-the-art transfer learning method for NLP. For the AG, Yelp, and DBpedia datasets, we compare against the state-of-the-art text categorization method by \newcite{johnson2017deep}.

\begin{figure*}[!htb]
    \begin{subfigure}{.31\linewidth}
      \centering
         \includegraphics[height=1.4in]{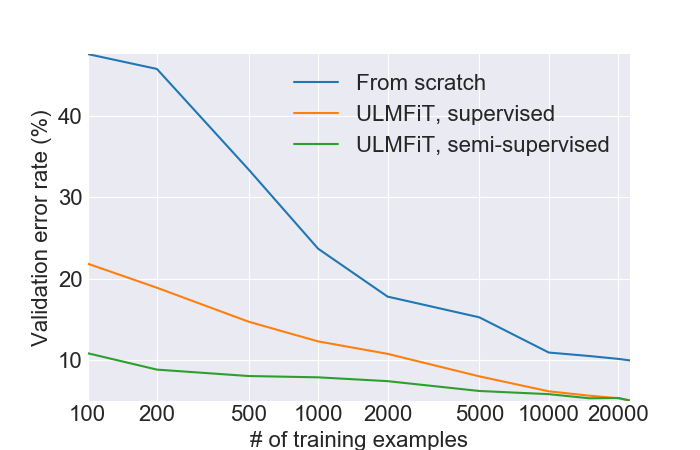}
    \end{subfigure}%
    \hspace*{0.3cm}
    \begin{subfigure}{.31\linewidth}
      \centering
         \includegraphics[height=1.4in]{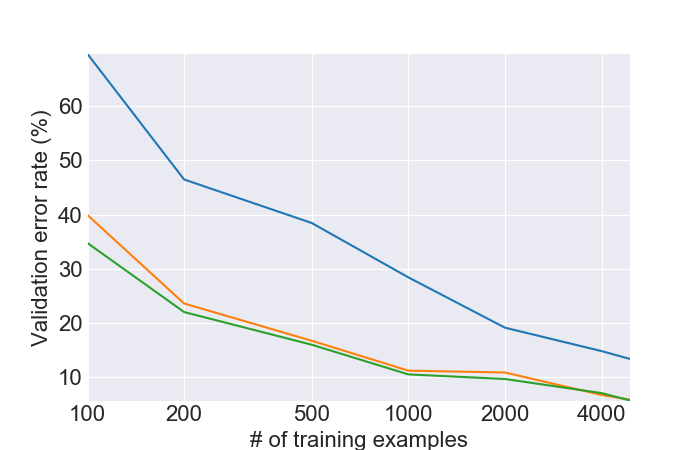}
    \end{subfigure}
    \hspace*{0.3cm}
    \begin{subfigure}{.31\linewidth}
      \centering
         \includegraphics[height=1.4in]{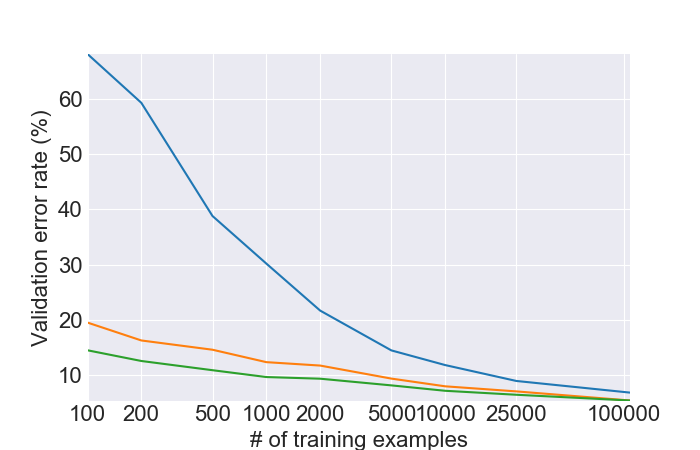}
    \end{subfigure}
    \caption{Validation error rates for supervised and semi-supervised ULMFiT vs. training from scratch with different numbers of training examples on IMDb, TREC-6, and AG (from left to right).}
\label{fig:few-shot_learning}
\end{figure*}

\subsection{Results}

For consistency, we report all results as error rates (lower is better). We show the test error rates on the IMDb and TREC-6 datasets used by \newcite{Mccann2017} in Table \ref{tab:results-mccann}. Our method outperforms both CoVe, a state-of-the-art transfer learning method based on hypercolumns, as well as the state-of-the-art on both datasets. On IMDb, we reduce the error dramatically by 43.9\% and 22\% with regard to CoVe and the state-of-the-art respectively. This is promising as the existing state-of-the-art requires complex architectures \cite{deepcontext2017}, multiple forms of attention \cite{Mccann2017} and sophisticated embedding schemes \cite{johnson2016supervised}, while our method employs a regular LSTM with dropout. We note that the language model fine-tuning approach of \newcite{Dai2015} only achieves an error of 7.64 vs. 4.6 for our method  on IMDb, demonstrating the benefit of transferring knowledge from a large ImageNet-like corpus using our fine-tuning techniques. IMDb in particular is reflective of real-world datasets: Its documents are generally a few paragraphs long---similar to emails (e.g for legal discovery) and online comments (e.g for community management); and sentiment analysis is similar to many commercial applications, e.g. product response tracking and support email routing.

On TREC-6, our improvement---similar as the improvements of state-of-the-art approaches---is not statistically significant, due to the small size of the 500-examples test set.
Nevertheless, the competitive performance on TREC-6 demonstrates that our model performs well across different dataset sizes and can deal with examples that range from single sentences---in the case of TREC-6---to several paragraphs for IMDb. Note that despite pretraining on more than two orders of magnitude less data than the 7 million sentence pairs used by \newcite{Mccann2017}, we consistently outperform their approach on both datasets.

We show the test error rates on the larger AG, DBpedia, Yelp-bi, and Yelp-full datasets in Table \ref{tab:results-zhang}. Our method again outperforms the state-of-the-art significantly. On AG, we observe a similarly dramatic error reduction by 23.7\% compared to the state-of-the-art. On DBpedia, Yelp-bi, and Yelp-full, we reduce the error by 4.8\%, 18.2\%, 2.0\% respectively.

\section{Analysis} \label{sec:analysis}

In order to assess the impact of each contribution, we perform a series of analyses and ablations. We run experiments on three corpora, IMDb, TREC-6, and AG that are representative of different tasks, genres, and sizes. For all experiments, we split off $10\%$ of the training set and report error rates on this validation set with unidirectional LMs. We fine-tune the classifier for $50$ epochs and train all methods but ULMFiT with early stopping.

\paragraph{Low-shot learning} One of the main benefits of transfer learning is being able to train a model for a task with a small number of labels. We evaluate ULMFiT on different numbers of labeled examples in two settings: only labeled examples are used for LM fine-tuning (`\emph{supervised}'); and all task data is available and can be used to fine-tune the LM (`\emph{semi-supervised}'). We compare ULMFiT to training from scratch---which is necessary for hypercolumn-based approaches. We split off balanced fractions of the training data, keep the validation set fixed, and use the same hyperparameters as before. We show the results in Figure \ref{fig:few-shot_learning}.

On IMDb and AG, supervised ULMFiT with only $100$ labeled examples matches the performance of training from scratch with $10\times$ and $20\times$ more data respectively, clearly demonstrating the benefit of general-domain LM pretraining. If we allow ULMFiT to also utilize unlabeled examples ($50$k for IMDb, $100$k for AG), at $100$ labeled examples, we match the performance of training from scratch with $50\times$ and $100\times$ more data on AG and IMDb respectively. On TREC-6, ULMFiT significantly improves upon training from scratch; as examples are shorter and fewer, supervised and semi-supervised ULMFiT achieve similar results.

\paragraph{Impact of pretraining} We compare using no pretraining with pretraining on WikiText-103 \cite{Merity2016} in Table \ref{tab:pretraining-corpus}. Pretraining is most useful for small and medium-sized datasets, which are most common in commercial applications. However, even for large datasets, pretraining improves performance.

\begin{table}
\centering
\begin{tabular}{l c c c}
\toprule
Pretraining & IMDb & TREC-6 & AG \\
 \midrule
Without pretraining & 5.63 & 10.67 & 5.52 \\
With pretraining & \textbf{5.00} & \textbf{5.69} & \textbf{5.38}\\
\bottomrule
\end{tabular}
\caption{Validation error rates for ULMFiT with and without pretraining.}
\label{tab:pretraining-corpus}
\end{table}

\begin{table}
\centering
\begin{tabular}{l c c c}
\toprule
LM & IMDb & TREC-6 & AG \\
 \midrule
Vanilla LM & 5.98 & 7.41 & 5.76 \\
AWD-LSTM LM & \textbf{5.00} & \textbf{5.69} & \textbf{5.38}\\
\bottomrule
\end{tabular}
\caption{Validation error rates for ULMFiT with a vanilla LM and the AWD-LSTM LM.}
\label{tab:lm-choice}
\end{table}

\paragraph{Impact of LM quality} In order to gauge the importance of choosing an appropriate LM, we compare a vanilla LM with the same hyperparameters without any dropout\footnote{To avoid overfitting, we only train the vanilla LM classifier for $5$ epochs and keep dropout of $0.4$ in the classifier.} with the AWD-LSTM LM with tuned dropout parameters in Table \ref{tab:lm-choice}. Using our fine-tuning techniques, even a regular LM reaches surprisingly good performance on the larger datasets. On the smaller TREC-6, a vanilla LM without dropout runs the risk of overfitting, which decreases performance. 

\paragraph{Impact of LM fine-tuning} We compare no fine-tuning against fine-tuning the full model \cite{erhan2010does} (`\emph{Full}'), the most commonly used fine-tuning method, with and without discriminative fine-tuning (`\emph{Discr}') and slanted triangular learning rates (`\emph{Stlr}') in Table \ref{tab:lm-fine-tuning}. Fine-tuning the LM is most beneficial for larger datasets. `\emph{Discr}' and `\emph{Stlr}' improve performance across all three datasets and are necessary on the smaller TREC-6, where regular fine-tuning is not beneficial.

\begin{table}
\centering
\begin{tabular}{l c c c}
\toprule
LM fine-tuning & IMDb & TREC-6 & AG\\
 \midrule
No LM fine-tuning & 6.99 & 6.38 & 6.09\\
Full & 5.86 & 6.54 & 5.61 \\
Full + discr & 5.55 & 6.36 & 5.47\\
Full + discr + stlr & \textbf{5.00} & \textbf{5.69} & \textbf{5.38} \\
\bottomrule
\end{tabular}
\caption{Validation error rates for ULMFiT with different variations of LM fine-tuning.}
\label{tab:lm-fine-tuning}
\end{table}

\paragraph{Impact of classifier fine-tuning} We compare training from scratch, fine-tuning the full model (`\emph{Full}'), only fine-tuning the last layer (`\emph{Last}') \cite{donahue2014decaf}, `\emph{Chain-thaw}' \cite{Felbo2017}, and gradual unfreezing (`\emph{Freez}'). We furthermore assess the importance of discriminative fine-tuning (`\emph{Discr}') and slanted triangular learning rates (`\emph{Stlr}'). We compare the latter to an alternative, aggressive cosine annealing schedule (`\emph{Cos}') \cite{Loshchilov2017}. We use a learning rate $\eta^L=0.01$ for `\emph{Discr}', learning rates of $0.001$ and $0.0001$ for the last and all other layers respectively for `\emph{Chain-thaw}' as in \cite{Felbo2017}, and a learning rate of $0.001$ otherwise. We show the results in Table \ref{tab:target-task-fine-tuning}. 

\begin{table}
\centering
\begin{tabular}{l c c c}
\toprule
Classifier fine-tuning & IMDb & TREC-6 & AG \\
 \midrule
From scratch & 9.93 & 13.36 & 6.81 \\
Full & 6.87 & 6.86 & 5.81 \\
Full + discr & 5.57 & 6.21 & 5.62\\
Last & 6.49 & 16.09 & 8.38 \\
Chain-thaw & 5.39 & 6.71 & 5.90\\
Freez & 6.37 & 6.86 & 5.81 \\
Freez + discr & 5.39 & 5.86 & 6.04 \\
Freez + stlr & 5.04 & 6.02 & 5.35 \\
Freez + cos & 5.70 & 6.38 & \textbf{5.29} \\
Freez + discr + stlr & \textbf{5.00} & \textbf{5.69} & 5.38 \\
\bottomrule
\end{tabular}
\caption{Validation error rates for ULMFiT with different methods to fine-tune the classifier.}
\label{tab:target-task-fine-tuning}
\end{table}

Fine-tuning the classifier significantly improves over training from scratch, particularly on the small TREC-6. `\emph{Last}', the standard fine-tuning method in CV, severely underfits and is never able to lower the training error to $0$. `\emph{Chain-thaw}' achieves competitive performance on the smaller datasets, but is outperformed significantly on the large AG. `\emph{Freez}' provides similar performance as `\emph{Full}'. `\emph{Discr}' consistently boosts the performance of `\emph{Full}' and `\emph{Freez}', except for the large AG. Cosine annealing is competitive with slanted triangular learning rates on large data, but under-performs on smaller datasets. Finally, full ULMFiT classifier fine-tuning (bottom row) achieves the best performance on IMDB and TREC-6 and competitive performance on AG. Importantly, ULMFiT is the only method that shows excellent performance across the board---and is therefore the only \emph{universal} method.

\begin{figure}[!htb]
    \begin{subfigure}{\linewidth}
      \centering
         \includegraphics[width=\linewidth]{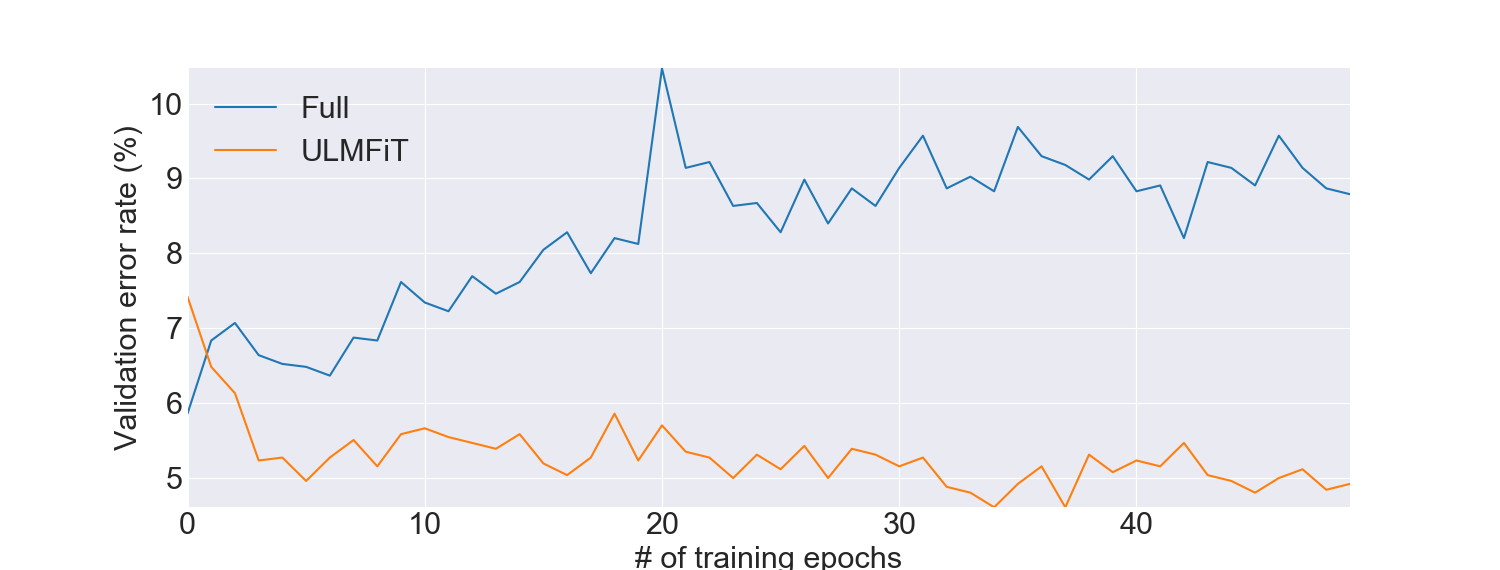}
    \end{subfigure}%
    \hspace{10cm}
    \begin{subfigure}{\linewidth}
      \centering
         \includegraphics[width=\linewidth]{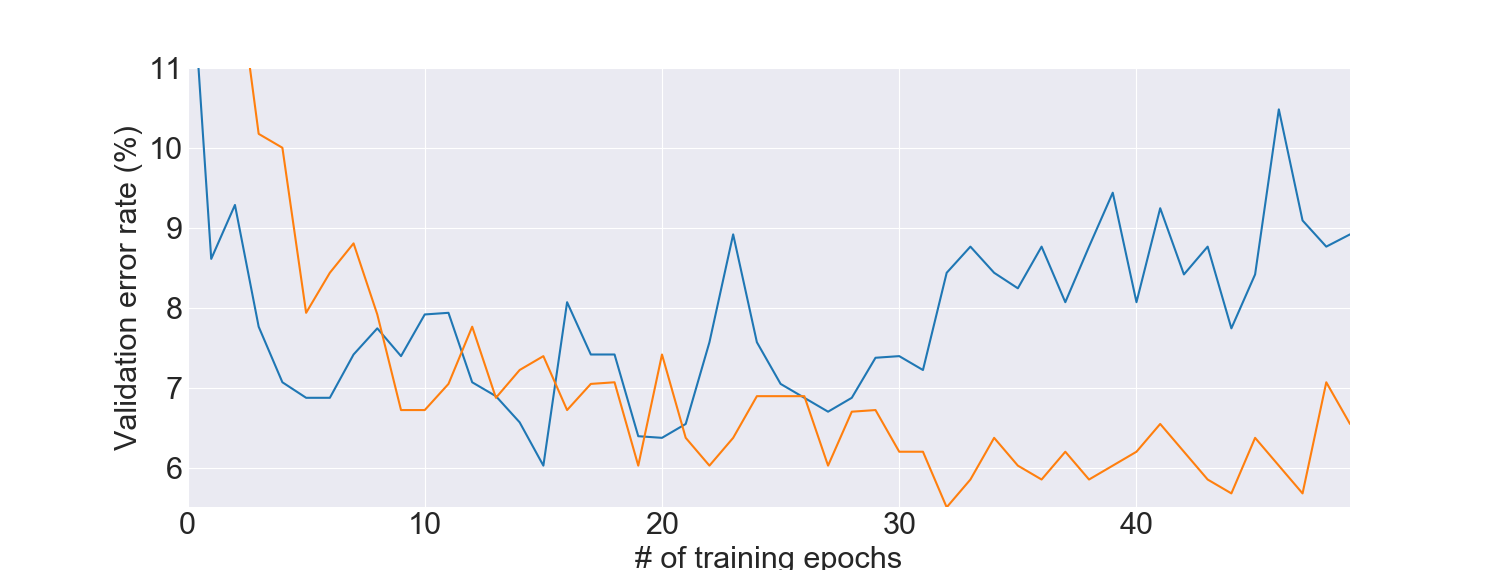}
    \end{subfigure}
    \begin{subfigure}{\linewidth}
      \centering
         \includegraphics[width=\linewidth]{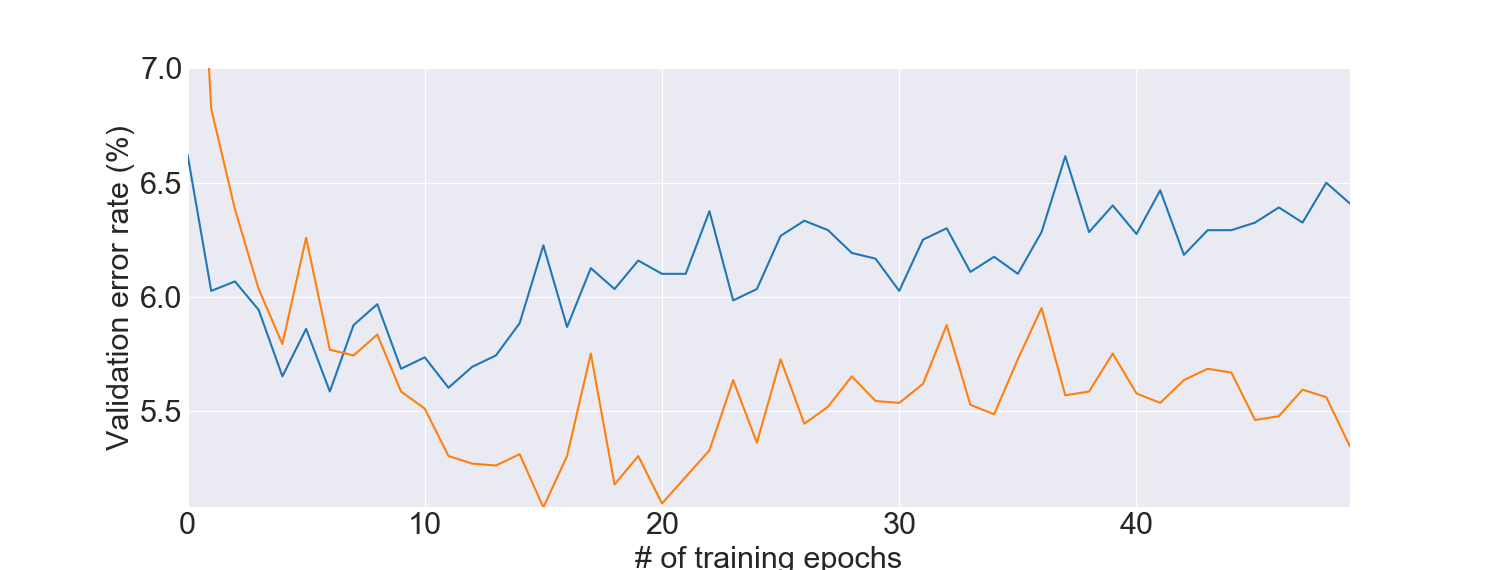}
    \end{subfigure}
    \caption{Validation error rate curves for fine-tuning the classifier with ULMFiT and `\emph{Full}' on IMDb, TREC-6, and AG (top to bottom).}
\label{fig:fine-tuning}
\end{figure}

\paragraph{Classifier fine-tuning behavior} While our results demonstrate that \emph{how} we fine-tune the classifier makes a significant difference, fine-tuning for inductive transfer is currently under-explored in NLP as it mostly has been thought to be unhelpful \cite{Mou2016}. To better understand the fine-tuning behavior of our model, we compare the validation error of the classifier fine-tuned with ULMFiT and `\emph{Full}' during training in Figure \ref{fig:fine-tuning}.

On all datasets, fine-tuning the full model leads to the lowest error comparatively early in training, e.g. already after the first epoch on IMDb. The error then increases as the model starts to overfit and knowledge captured through pretraining is lost. In contrast, ULMFiT is more stable and suffers from no such catastrophic forgetting; performance remains similar or improves until late epochs, which shows the positive effect of the learning rate schedule.

\paragraph{Impact of bidirectionality} At the cost of training a second model, ensembling the predictions of a forward and backwards LM-classifier brings a performance boost of around $0.5$--$0.7$. On IMDb we lower the test error from $5.30$ of a single model to $4.58$ for the bidirectional model.

\section{Discussion and future directions}

While we have shown that ULMFiT can achieve state-of-the-art performance on widely used text classification tasks, we believe that language model fine-tuning will be particularly useful in the following settings compared to existing transfer learning approaches \cite{Conneau2017,Mccann2017,deepcontext2017}: a) NLP for non-English languages, where training data for supervised pretraining tasks is scarce; b) new NLP tasks where no state-of-the-art architecture exists; and c) tasks with limited amounts of labeled data (and some amounts of unlabeled data).

Given that transfer learning and particularly fine-tuning for NLP is under-explored, many future directions are possible. One possible direction is to improve language model pretraining and fine-tuning and make them more scalable: for ImageNet, predicting far fewer classes only incurs a small performance drop \cite{Huh2016}, while recent work shows that an alignment between source and target task label sets is important \cite{Mahajan2018}---focusing on predicting a subset of words such as the most frequent ones might retain most of the performance while speeding up training. Language modeling can also be augmented with additional tasks in a multi-task learning fashion \cite{Caruana1993} or enriched with additional supervision, e.g. syntax-sensitive dependencies \cite{linzen2016assessing} to create a model that is more general or better suited for certain downstream tasks, ideally in a weakly-supervised manner to retain its universal properties.

Another direction is to apply the method to novel tasks and models. While an extension to sequence labeling is straightforward, other tasks with more complex interactions such as entailment or question answering may require novel ways to pretrain and fine-tune. Finally, while we have provided a series of analyses and ablations, more studies are required to better understand what knowledge a pretrained language model captures, how this changes during fine-tuning, and what information different tasks require.

\section{Conclusion}

We have proposed ULMFiT, an effective and extremely sample-efficient transfer learning method that can be applied to any NLP task. We have also proposed several novel fine-tuning techniques that in conjunction prevent catastrophic forgetting and enable robust learning across a diverse range of tasks. Our method significantly outperformed existing transfer learning techniques and the state-of-the-art on six representative text classification tasks. We hope that our results will catalyze new developments in transfer learning for NLP.

\section*{Acknowledgments}

We thank the anonymous reviewers for their valuable feedback. Sebastian is supported by Irish Research Council Grant Number EBPPG/2014/30 and Science Foundation Ireland Grant Number SFI/12/RC/2289.

\bibliography{pre-training}

\begin{thebibliography}{}
\expandafter\ifx\csname natexlab\endcsname\relax\def\natexlab#1{#1}\fi

\bibitem[{Baxter(2000)}]{Baxter2000}
Jonathan Baxter. 2000.
\newblock {A Model of Inductive Bias Learning}.
\newblock {\em Journal of Artificial Intelligence Research\/} 12:149--198.

\bibitem[{Blitzer et~al.(2007)Blitzer, Dredze, and Pereira}]{Blitzer2007}
John Blitzer, Mark Dredze, and Fernando Pereira. 2007.
\newblock \href{https://doi.org/10.1109/IRPS.2011.5784441}{{Biographies,
  bollywood, boom-boxes and blenders: Domain adaptation for sentiment
  classification}}.
\newblock {\em Annual Meeting-Association for Computational Linguistics\/}
  45(1):440.
\newblock
  \href{https://doi.org/10.1109/IRPS.2011.5784441}{https://doi.org/10.1109/IRPS.2011.5784441}.

\bibitem[{Caragea et~al.(2011)Caragea, McNeese, Jaiswal, Traylor, Kim, Mitra,
  Wu, Tapia, Giles, Jansen et~al.}]{caragea2011classifying}
Cornelia Caragea, Nathan McNeese, Anuj Jaiswal, Greg Traylor, Hyun-Woo Kim,
  Prasenjit Mitra, Dinghao Wu, Andrea~H Tapia, Lee Giles, Bernard~J Jansen,
  et~al. 2011.
\newblock Classifying text messages for the haiti earthquake.
\newblock In {\em Proceedings of the 8th international conference on
  information systems for crisis response and management (ISCRAM2011)\/}.
  Citeseer.

\bibitem[{Caruana(1993)}]{Caruana1993}
Rich Caruana. 1993.
\newblock {Multitask learning: A knowledge-based source of inductive bias}.
\newblock In {\em Proceedings of the Tenth International Conference on Machine
  Learning\/}.

\bibitem[{Chen et~al.(2017)Chen, Badrinarayanan, Lee, and
  Rabinovich}]{Chen2017}
Zhao Chen, Vijay Badrinarayanan, Chen-Yu Lee, and Andrew Rabinovich. 2017.
\newblock {GradNorm: Gradient Normalization for Adaptive Loss Balancing in Deep
  Multitask Networks} pages 1--10.

\bibitem[{Chu et~al.(2012)Chu, Gianvecchio, Wang, and
  Jajodia}]{chu2012detecting}
Zi~Chu, Steven Gianvecchio, Haining Wang, and Sushil Jajodia. 2012.
\newblock Detecting automation of twitter accounts: Are you a human, bot, or
  cyborg?
\newblock {\em IEEE Transactions on Dependable and Secure Computing\/}
  9(6):811--824.

\bibitem[{Conneau et~al.(2017)Conneau, Kiela, Schwenk, Barrault, and
  Bordes}]{Conneau2017}
Alexis Conneau, Douwe Kiela, Holger Schwenk, Lo{\"{i}}c Barrault, and Antoine
  Bordes. 2017.
\newblock {Supervised Learning of Universal Sentence Representations from
  Natural Language Inference Data}.
\newblock In {\em Proceedings of the 2017 Conference on Empirical Methods in
  Natural Language Processing\/}.

\bibitem[{Dai and Le(2015)}]{Dai2015}
Andrew~M. Dai and Quoc~V. Le. 2015.
\newblock \href{http://arxiv.org/abs/1511.01432}{{Semi-supervised Sequence
  Learning}}.
\newblock {\em Advances in Neural Information Processing Systems (NIPS '15)\/}
  \href{http://arxiv.org/abs/1511.01432}{http://arxiv.org/abs/1511.01432}.

\bibitem[{Donahue et~al.(2014)Donahue, Jia, Vinyals, Hoffman, Zhang, Tzeng, and
  Darrell}]{donahue2014decaf}
Jeff Donahue, Yangqing Jia, Oriol Vinyals, Judy Hoffman, Ning Zhang, Eric
  Tzeng, and Trevor Darrell. 2014.
\newblock Decaf: A deep convolutional activation feature for generic visual
  recognition.
\newblock In {\em International conference on machine learning\/}. pages
  647--655.

\bibitem[{Dozat and Manning(2017)}]{Dozat2017}
Timothy Dozat and Christopher~D. Manning. 2017.
\newblock {Deep Biaffine Attention for Neural Dependency Parsing}.
\newblock In {\em Proceedings of ICLR 2017\/}.

\bibitem[{Erhan et~al.(2010)Erhan, Bengio, Courville, Manzagol, Vincent, and
  Bengio}]{erhan2010does}
Dumitru Erhan, Yoshua Bengio, Aaron Courville, Pierre-Antoine Manzagol, Pascal
  Vincent, and Samy Bengio. 2010.
\newblock Why does unsupervised pre-training help deep learning?
\newblock {\em Journal of Machine Learning Research\/} 11(Feb):625--660.

\bibitem[{Felbo et~al.(2017)Felbo, Mislove, S{\o}gaard, Rahwan, and
  Lehmann}]{Felbo2017}
Bjarke Felbo, Alan Mislove, Anders S{\o}gaard, Iyad Rahwan, and Sune Lehmann.
  2017.
\newblock {Using millions of emoji occurrences to learn any-domain
  representations for detecting sentiment, emotion and sarcasm}.
\newblock In {\em Proceedings of the 2017 Conference on Empirical Methods in
  Natural Language Processing\/}.

\bibitem[{Gulordava et~al.(2018)Gulordava, Bojanowski, Grave, Linzen, and
  Baroni}]{Gulordava2018}
Kristina Gulordava, Piotr Bojanowski, Edouard Grave, Tal Linzen, and Marco
  Baroni. 2018.
\newblock {Colorless green recurrent networks dream hierarchically}.
\newblock In {\em Proceedings of NAACL-HLT 2018\/}.

\bibitem[{Hariharan et~al.(2015)Hariharan, Arbel{\'a}ez, Girshick, and
  Malik}]{hariharan2015hypercolumns}
Bharath Hariharan, Pablo Arbel{\'a}ez, Ross Girshick, and Jitendra Malik. 2015.
\newblock Hypercolumns for object segmentation and fine-grained localization.
\newblock In {\em Proceedings of the IEEE Conference on Computer Vision and
  Pattern Recognition\/}. pages 447--456.

\bibitem[{He et~al.(2016)He, Zhang, Ren, and Sun}]{He2015}
Kaiming He, Xiangyu Zhang, Shaoqing Ren, and Jian Sun. 2016.
\newblock {Deep Residual Learning for Image Recognition}.
\newblock In {\em Proceedings of the IEEE Conference on Computer Vision and
  Pattern Recognition\/}.

\bibitem[{Huang et~al.(2017)Huang, Liu, Weinberger, and van~der
  Maaten}]{Huang2017}
Gao Huang, Zhuang Liu, Kilian~Q. Weinberger, and Laurens van~der Maaten. 2017.
\newblock {Densely Connected Convolutional Networks}.
\newblock In {\em Proceedings of CVPR 2017\/}.

\bibitem[{Huh et~al.(2016)Huh, Agrawal, and Efros}]{Huh2016}
Minyoung Huh, Pulkit Agrawal, and Alexei~A Efros. 2016.
\newblock {What makes ImageNet good for transfer learning?}
\newblock {\em arXiv preprint arXiv:1608.08614\/} .

\bibitem[{Ioffe and Szegedy(2015)}]{ioffe2015batch}
Sergey Ioffe and Christian Szegedy. 2015.
\newblock Batch normalization: Accelerating deep network training by reducing
  internal covariate shift.
\newblock In {\em International Conference on Machine Learning\/}. pages
  448--456.

\bibitem[{Jindal and Liu(2007)}]{jindal2007review}
Nitin Jindal and Bing Liu. 2007.
\newblock Review spam detection.
\newblock In {\em Proceedings of the 16th international conference on World
  Wide Web\/}. ACM, pages 1189--1190.

\bibitem[{Johnson and Zhang(2016)}]{johnson2016supervised}
Rie Johnson and Tong Zhang. 2016.
\newblock Supervised and semi-supervised text categorization using lstm for
  region embeddings.
\newblock In {\em International Conference on Machine Learning\/}. pages
  526--534.

\bibitem[{Johnson and Zhang(2017)}]{johnson2017deep}
Rie Johnson and Tong Zhang. 2017.
\newblock Deep pyramid convolutional neural networks for text categorization.
\newblock In {\em Proceedings of the 55th Annual Meeting of the Association for
  Computational Linguistics (Volume 1: Long Papers)\/}. volume~1, pages
  562--570.

\bibitem[{Linzen et~al.(2016)Linzen, Dupoux, and
  Goldberg}]{linzen2016assessing}
Tal Linzen, Emmanuel Dupoux, and Yoav Goldberg. 2016.
\newblock Assessing the ability of lstms to learn syntax-sensitive
  dependencies.
\newblock {\em arXiv preprint arXiv:1611.01368\/} .

\bibitem[{Liu et~al.(2018)Liu, Shang, Xu, Ren, Gui, Peng, and
  Han}]{empower2018liu}
Liyuan Liu, Jingbo Shang, Frank Xu, Xiang Ren, Huan Gui, Jian Peng, and Jiawei
  Han. 2018.
\newblock Empower sequence labeling with task-aware neural language model.
\newblock In {\em Proceedings of AAAI 2018\/}.

\bibitem[{Long et~al.(2015{\natexlab{a}})Long, Shelhamer, and
  Darrell}]{long2015fully}
Jonathan Long, Evan Shelhamer, and Trevor Darrell. 2015{\natexlab{a}}.
\newblock Fully convolutional networks for semantic segmentation.
\newblock In {\em Proceedings of the IEEE Conference on Computer Vision and
  Pattern Recognition\/}. pages 3431--3440.

\bibitem[{Long et~al.(2015{\natexlab{b}})Long, Cao, Wang, and
  Jordan}]{Long2015a}
Mingsheng Long, Yue Cao, Jianmin Wang, and Michael~I. Jordan.
  2015{\natexlab{b}}.
\newblock {Learning Transferable Features with Deep Adaptation Networks}.
\newblock In {\em Proceedings of the 32nd International Conference on Machine
  learning (ICML '15)\/}. volume~37.

\bibitem[{Loshchilov and Hutter(2017)}]{Loshchilov2017}
Ilya Loshchilov and Frank Hutter. 2017.
\newblock {SGDR: Stochastic Gradient Descent with Warm Restarts}.
\newblock In {\em Proceedings of the Internal Conference on Learning
  Representations 2017\/}.

\bibitem[{Maas et~al.(2011)Maas, Daly, Pham, Huang, Ng, and
  Potts}]{maas2011learning}
Andrew~L Maas, Raymond~E Daly, Peter~T Pham, Dan Huang, Andrew~Y Ng, and
  Christopher Potts. 2011.
\newblock Learning word vectors for sentiment analysis.
\newblock In {\em Proceedings of the 49th Annual Meeting of the Association for
  Computational Linguistics: Human Language Technologies-Volume 1\/}.
  Association for Computational Linguistics, pages 142--150.

\bibitem[{Mahajan et~al.(2018)Mahajan, Girshick, Ramanathan, He, Paluri, Li,
  Bharambe, and van~der Maaten}]{Mahajan2018}
Dhruv Mahajan, Ross Girshick, Vignesh Ramanathan, Kaiming He, Manohar Paluri,
  Yixuan Li, Ashwin Bharambe, and Laurens van~der Maaten. 2018.
\newblock {Exploring the Limits of Weakly Supervised Pretraining} .

\bibitem[{McCann et~al.(2017)McCann, Bradbury, Xiong, and Socher}]{Mccann2017}
Bryan McCann, James Bradbury, Caiming Xiong, and Richard Socher. 2017.
\newblock {Learned in Translation: Contextualized Word Vectors}.
\newblock In {\em Advances in Neural Information Processing Systems\/}.

\bibitem[{Merity et~al.(2017{\natexlab{a}})Merity, {Shirish Keskar}, and
  Socher}]{Merity2017}
Stephen Merity, Nitish {Shirish Keskar}, and Richard Socher.
  2017{\natexlab{a}}.
\newblock {Regularizing and Optimizing LSTM Language Models}.
\newblock {\em arXiv preprint arXiv:1708.02182\/} .

\bibitem[{Merity et~al.(2017{\natexlab{b}})Merity, Xiong, Bradbury, and
  Socher}]{Merity2016}
Stephen Merity, Caiming Xiong, James Bradbury, and Richard Socher.
  2017{\natexlab{b}}.
\newblock {Pointer Sentinel Mixture Models}.
\newblock In {\em Proceedings of the International Conference on Learning
  Representations 2017\/}.

\bibitem[{Mikolov et~al.(2013)Mikolov, Chen, Corrado, and Dean}]{Mikolov2013d}
Tomas Mikolov, Kai Chen, Greg Corrado, and Jeffrey Dean. 2013.
\newblock {Distributed Representations of Words and Phrases and their
  Compositionality}.
\newblock In {\em Advances in Neural Information Processing Systems\/}.

\bibitem[{Min et~al.(2017)Min, Seo, and Hajishirzi}]{Min2017}
Sewon Min, Minjoon Seo, and Hannaneh Hajishirzi. 2017.
\newblock {Question Answering through Transfer Learning from Large Fine-grained
  Supervision Data}.
\newblock In {\em Proceedings of the 55th Annual Meeting of the Association for
  Computational Linguistics (Short Papers)\/}.

\bibitem[{Miyato et~al.(2016)Miyato, Dai, and
  Goodfellow}]{miyato2016adversarial}
Takeru Miyato, Andrew~M Dai, and Ian Goodfellow. 2016.
\newblock Adversarial training methods for semi-supervised text classification.
\newblock {\em arXiv preprint arXiv:1605.07725\/} .

\bibitem[{Mou et~al.(2016)Mou, Meng, Yan, Li, Xu, Zhang, and Jin}]{Mou2016}
Lili Mou, Zhao Meng, Rui Yan, Ge~Li, Yan Xu, Lu~Zhang, and Zhi Jin. 2016.
\newblock {How Transferable are Neural Networks in NLP Applications?}
\newblock {\em Proceedings of 2016 Conference on Empirical Methods in Natural
  Language Processing\/} .

\bibitem[{Mou et~al.(2015)Mou, Peng, Li, Xu, Zhang, and
  Jin}]{mou2015discriminative}
Lili Mou, Hao Peng, Ge~Li, Yan Xu, Lu~Zhang, and Zhi Jin. 2015.
\newblock Discriminative neural sentence modeling by tree-based convolution.
\newblock In {\em Proceedings of the 2015 Conference on Empirical Methods in
  Natural Language Processing\/}.

\bibitem[{Ngai et~al.(2011)Ngai, Hu, Wong, Chen, and Sun}]{ngai2011application}
EWT Ngai, Yong Hu, YH~Wong, Yijun Chen, and Xin Sun. 2011.
\newblock The application of data mining techniques in financial fraud
  detection: A classification framework and an academic review of literature.
\newblock {\em Decision Support Systems\/} 50(3):559--569.

\bibitem[{Pan and Yang(2010)}]{Pan2010}
Sinno~Jialin Pan and Qiang Yang. 2010.
\newblock {A survey on transfer learning}.
\newblock {\em IEEE Transactions on Knowledge and Data Engineering\/}
  22(10):1345--1359.

\bibitem[{Peters et~al.(2017)Peters, Ammar, Bhagavatula, and
  Power}]{peters2017semi}
Matthew~E Peters, Waleed Ammar, Chandra Bhagavatula, and Russell Power. 2017.
\newblock Semi-supervised sequence tagging with bidirectional language models.
\newblock In {\em Proceedings of ACL 2017\/}.

\bibitem[{Peters et~al.(2018)Peters, Neumann, Iyyer, Gardner, Clark, Lee, and
  Zettlemoyer}]{deepcontext2017}
Matthew~E Peters, Mark Neumann, Mohit Iyyer, Matt Gardner, Christopher Clark,
  Kenton Lee, and Luke Zettlemoyer. 2018.
\newblock Deep contextualized word representations.
\newblock In {\em Proceedings of NAACL 2018\/}.

\bibitem[{Radford et~al.(2017)Radford, Jozefowicz, and
  Sutskever}]{radford2017learning}
Alec Radford, Rafal Jozefowicz, and Ilya Sutskever. 2017.
\newblock Learning to generate reviews and discovering sentiment.
\newblock {\em arXiv preprint arXiv:1704.01444\/} .

\bibitem[{Rei(2017)}]{semisupervised2017rei}
Marek Rei. 2017.
\newblock Semi-supervised multitask learning for sequence labeling.
\newblock In {\em Proceedings of ACL 2017\/}.

\bibitem[{Roitblat et~al.(2010)Roitblat, Kershaw, and
  Oot}]{roitblat2010document}
Herbert~L Roitblat, Anne Kershaw, and Patrick Oot. 2010.
\newblock Document categorization in legal electronic discovery: computer
  classification vs. manual review.
\newblock {\em Journal of the Association for Information Science and
  Technology\/} 61(1):70--80.

\bibitem[{Ruder(2016)}]{ruder2016overview}
Sebastian Ruder. 2016.
\newblock An overview of gradient descent optimization algorithms.
\newblock {\em arXiv preprint arXiv:1609.04747\/} .

\bibitem[{Salakhutdinov and Hinton(2009)}]{salakhutdinov2009deep}
Ruslan Salakhutdinov and Geoffrey Hinton. 2009.
\newblock Deep boltzmann machines.
\newblock In {\em Artificial Intelligence and Statistics\/}. pages 448--455.

\bibitem[{Sennrich et~al.(2015)Sennrich, Haddow, and
  Birch}]{sennrich2015improving}
Rico Sennrich, Barry Haddow, and Alexandra Birch. 2015.
\newblock Improving neural machine translation models with monolingual data.
\newblock {\em arXiv preprint arXiv:1511.06709\/} .

\bibitem[{Severyn and Moschitti(2015)}]{Severyn2015a}
Aliaksei Severyn and Alessandro Moschitti. 2015.
\newblock {UNITN: Training Deep Convolutional Neural Network for Twitter
  Sentiment Classification}.
\newblock {\em Proceedings of the 9th International Workshop on Semantic
  Evaluation (SemEval 2015)\/} pages 464--469.

\bibitem[{Sharif~Razavian et~al.(2014)Sharif~Razavian, Azizpour, Sullivan, and
  Carlsson}]{sharif2014cnn}
Ali Sharif~Razavian, Hossein Azizpour, Josephine Sullivan, and Stefan Carlsson.
  2014.
\newblock Cnn features off-the-shelf: an astounding baseline for recognition.
\newblock In {\em Proceedings of the IEEE conference on computer vision and
  pattern recognition\/}. pages 806--813.

\bibitem[{Smith(2017)}]{smith2017cyclical}
Leslie~N Smith. 2017.
\newblock Cyclical learning rates for training neural networks.
\newblock In {\em Applications of Computer Vision (WACV), 2017 IEEE Winter
  Conference on\/}. IEEE, pages 464--472.

\bibitem[{Vapnik and Kotz(1982)}]{vapnik1982estimation}
Vladimir~Naumovich Vapnik and Samuel Kotz. 1982.
\newblock {\em Estimation of dependences based on empirical data\/}, volume~40.
\newblock Springer-Verlag New York.

\bibitem[{Voorhees and Tice(1999)}]{voorhees1999trec}
Ellen~M Voorhees and Dawn~M Tice. 1999.
\newblock The trec-8 question answering track evaluation.
\newblock In {\em TREC\/}. volume 1999, page~82.

\bibitem[{Wieting and Gimpel(2017)}]{Wieting2017}
John Wieting and Kevin Gimpel. 2017.
\newblock {Revisiting Recurrent Networks for Paraphrastic Sentence Embeddings}.
\newblock In {\em Proceedings of the 55th Annual Meeting of the Association for
  Computational Linguistics (ACL 2017)\/}.

\bibitem[{Yosinski et~al.(2014)Yosinski, Clune, Bengio, and
  Lipson}]{yosinski2014transferable}
Jason Yosinski, Jeff Clune, Yoshua Bengio, and Hod Lipson. 2014.
\newblock How transferable are features in deep neural networks?
\newblock In {\em Advances in neural information processing systems\/}. pages
  3320--3328.

\bibitem[{Zhang et~al.(2015)Zhang, Zhao, and LeCun}]{zhang2015character}
Xiang Zhang, Junbo Zhao, and Yann LeCun. 2015.
\newblock Character-level convolutional networks for text classification.
\newblock In {\em Advances in neural information processing systems\/}. pages
  649--657.

\bibitem[{Zhou et~al.(2016)Zhou, Qi, Zheng, Xu, Bao, and Xu}]{zhou2016text}
Peng Zhou, Zhenyu Qi, Suncong Zheng, Jiaming Xu, Hongyun Bao, and Bo~Xu. 2016.
\newblock Text classification improved by integrating bidirectional lstm with
  two-dimensional max pooling.
\newblock In {\em Proceedings of COLING 2016\/}.

\end{thebibliography}
\bibliographystyle{acl_natbib}

\end{document}